\documentclass{article}

\PassOptionsToPackage{numbers,sort&compress}{natbib}

\usepackage[preprint]{neurips_2026}

\workshoptitle{AgenticOS: Co-designing Systems and ML Foundations of an OS Layer for Agentic AI}

\usepackage[utf8]{inputenc}
\usepackage[T1]{fontenc}
\usepackage{courier}
\usepackage{amsmath,amssymb,amsthm}
\usepackage{booktabs}
\usepackage{graphicx}
\usepackage{placeins}
\usepackage{microtype}
\usepackage{xcolor}
\usepackage{xurl}
\usepackage{tikz}
\usetikzlibrary{arrows.meta,positioning}
\usepackage[hidelinks]{hyperref}

\definecolor{navy}{HTML}{173B57}
\definecolor{teal}{HTML}{167A78}
\definecolor{paleBlue}{HTML}{EAF2F8}
\definecolor{paleTeal}{HTML}{E7F4F2}

\newcommand{\method}{\textsc{LiveLoRA}}
\newcommand{\R}{\mathbb{R}}
\newcommand{\cache}{\mathcal{C}}
\newcommand{\state}{\mathcal{S}}
\newcommand{\model}{\mathcal{M}}
\newcommand{\KL}{\operatorname{KL}}
\newtheorem{theorem}{Theorem}

\title{LiveLoRA: Exact KV Cache Migration Across LoRA Updates for LLM Agents}

\author{%
  Tianyi Huang \\
  Ryquo
  \And
  Samuel Xu \\
  App-In Club
  \And
  Ivy Gu \\
  App-In Club
  \And
  Nathan Huang \\
  App-In Club
  \And
  Siqi Zhang \\
  App-In Club
  \AND
  \normalfont
  \texttt{tianyi@ryquo.com}\quad
  \texttt{\{samuel, ivy, nathan.huang, siqi\}@appinclub.org}
}

\begin{document}

\maketitle

\begin{abstract}
Updating an LLM adapter should not require replaying the full interaction history of every active session. Yet reusing key--value (KV) cache entries computed under an earlier version mixes model versions, while full-history re-prefill becomes increasingly costly as histories and session fleets grow. We introduce \method{}, a model--runtime co-design for exact final-block KV cache migration. Frozen lower decoder blocks and fixed K/V adapter down-projections preserve each token's base projections and low-rank coordinates across adapter versions. The runtime reconstructs candidate K/V, reevaluates the latest token through the final block, and commits the adapter, cache, and next-token logits together. We prove continuation equivalence to full-history re-prefill in exact arithmetic under the stated model assumptions. On Qwen2.5-3B-Instruct, median paired publication speedups over full-history re-prefill average $68.98\times$ across five runs for 32 active sessions with 8K tokens each. Serving benchmarks show approximately 4.8\% decoding overhead relative to native cached decoding. Compared with retained-boundary replay, compact retention gives a $3.42\times$ reduction in additional session storage at 8K, with similar measured publication-plus-decoding cost. \method{} allows persistent LLM agents to adopt new adapter versions without replaying their full interaction histories.
\end{abstract}

\section{Introduction}

Persistent LLM agents maintain both a visible interaction history and hidden inference state. A low-rank adaptation (LoRA) update replaces trainable low-rank factors \citep{hu2022lora}, changing the parameters that produced the current KV cache. Continuing with that cache mixes model versions, whereas restoring coherence by replaying the session history incurs full-history re-prefill. This cost grows with context length and the number of active sessions and can dominate publication latency. This problem arises in online agent learning: agents can update an LLM policy from interaction \citep{carta2023glam}, recent parametric-memory agents apply LoRA updates within a rollout \citep{ren2026tmem}, and dynamic evaluation combines online LoRA with streaming KV caches \citep{rannentriki2024dynamic}. \method{} addresses how a new adapter version is applied to active sessions, independently of the learning rule.

Existing work factors adapter-dependent KV state across LoRA adapters, reconstructs cache state from retained activations, selectively recomputes state across fine-tuned variants, or invalidates stale entries around online updates \citep{jeon2026lragent,wang2026forkkv,gao2025hcache,liu2026droidspeak,liu2026local}. These approaches establish the value of reuse across models and versions. \method{} uses invariant coordinates to migrate active sessions across adapter versions: under explicit structural assumptions, it preserves the updated decoder's conditional distribution without replaying the frozen lower blocks and publishes the corresponding cache and logits together.

\method{} realizes this transition by coupling the adapter structure to the state retained by the runtime. Frozen lower blocks preserve the final-block input across versions; fixed K/V adapter down-projections turn that input into compact invariant coordinates. Immutable base projections and these coordinates suffice to rematerialize historical K/V, after which one latest-token computation supplies candidate logits. Retaining normalized inputs also supports trainable down-projections, trading memory for a broader adapter family. The resulting operation fits the execution layer of an operating system for agents: a parameter version becomes visible together with the inference state needed to execute it.

We make three contributions:
\begin{itemize}
    \item We formalize adapter publication as a version-coherent transition of adapter parameters, the KV cache, and next-token logits. For fixed-$A$ final-block K/V adaptation, we derive a closed-form migration and prove, under explicit assumptions, continuation equivalence to full-history re-prefill in exact arithmetic.
    \item We implement a process-local publication protocol with failure atomicity before commit: it prepares candidate state without mutating the currently published version, then commits the adapter, KV cache, and next-token logits as one coherent version.
    \item We characterize publication scaling and the latency--memory tradeoff against both full-history and retained-boundary replay. Serving benchmarks with growing histories apply the same coordinator safeguards to each strategy and measure publication latency and ordinary decoding cost separately.
\end{itemize}

\begin{figure}[t]
    \centering
    \includegraphics[width=\linewidth]{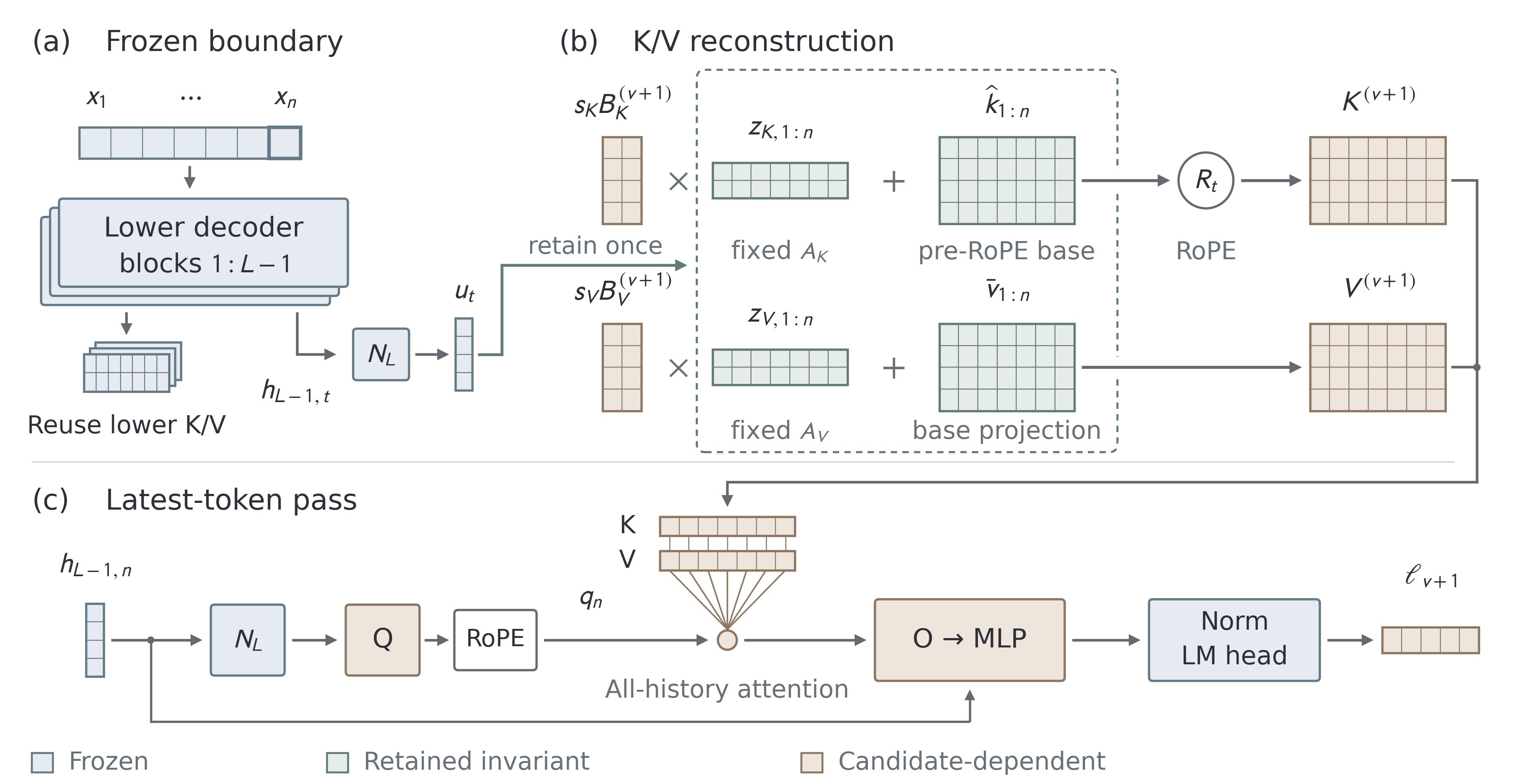}
    \caption{Fixed-$A$ final-block cache migration.
    (a) Retain invariant base projections and low-rank coordinates.
    (b) Reconstruct candidate K/V.
    (c) Reevaluate the latest token against the rebuilt history to obtain new logits.
    Lower-block caches are reused; the O-to-MLP stage includes residual connections and normalization.}
    \label{fig:migration}
\end{figure}

\section{Version-coherent state}
\label{sec:problem}

Let $\model_{\theta,\phi_v}$ be a decoder-only Transformer with frozen base parameters $\theta$ and adapter version $\phi_v$. For a fixed history $x_{1:n}$, complete prefill under $\phi_v$ defines the reference state
\begin{equation}
    \state_v^\star(x_{1:n})=\bigl(\phi_v,\cache_v(x_{1:n}),\ell_v(x_{1:n})\bigr).
    \label{eq:valid-state}
\end{equation}
Here $\cache_v$ is the KV cache and $\ell_v$ denotes next-token logits. A runtime state is \emph{version coherent} when its adapter, cache, and logits all correspond to one version; this excludes old--new mixtures without requiring bitwise agreement between different floating-point kernels. The exact-arithmetic target of migration is $\state_{v+1}^\star(x_{1:n})$ without changing the visible history. Section~\ref{sec:results} compares finite-precision serving with independent reference executions. We define \emph{publication latency} as the interval from when a candidate adapter becomes available until its next-token logits are ready. Training time is outside this interval because it is common to migration and full-history re-prefill.

\section{LiveLoRA}
\label{sec:method}

\subsection{Final-block adapter structure}

Consider a pre-norm decoder with $L$ blocks. \method{} places LoRA adapters in the final decoder block; embeddings, blocks $1{:}L-1$, all normalization layers, and the language-model head remain frozen. For position $t$,
\begin{equation}
    u_t=N_L(h_{L-1,t})\in\R^d.
    \label{eq:final-block-input}
\end{equation}
Here $h_{L-1,t}$ is the hidden state after block $L-1$, and $N_L$ normalizes it before final-block attention. With the history, attention mask, and positions fixed, $u_t$ is invariant across adapter versions. Final-block query (Q), attention output (O), and multilayer perceptron (MLP) projections use standard LoRA. Final-block K/V adapters fix the down-projection matrices, following LoRA-FA \citep{zhang2023lorafa}. For keys, let $W_K\in\R^{d_K\times d}$, $A_K\in\R^{r_K\times d}$, and $B_K^{(v)}\in\R^{d_K\times r_K}$ denote the base projection, fixed adapter down-projection, and versioned adapter up-projection, respectively, with frozen bias $b_K\in\R^{d_K}$ and scale $s_K$. Define $W_V\in\R^{d_V\times d}$, $A_V\in\R^{r_V\times d}$, $B_V^{(v)}\in\R^{d_V\times r_V}$, $b_V\in\R^{d_V}$, and $s_V$ analogously. Rotary position embedding (RoPE) is represented by a position-dependent linear map $R_t:\R^{d_K}\!\to\R^{d_K}$ that acts independently on each head \citep{su2024roformer}. Then
\begin{align}
    k_t^{(v)} &=R_t\!\left(W_Ku_t+b_K+s_KB_K^{(v)}A_Ku_t\right),
    \label{eq:key-version}\\
    v_t^{(v)} &=W_Vu_t+b_V+s_VB_V^{(v)}A_Vu_t.
    \label{eq:value-version}
\end{align}
Both $A_K$ and $A_V$ are fixed. Base weights, biases, ranks, scales, normalization, and positional parameters do not change across versions.

\subsection{Closed-form migration}

On the first evaluation of token $t$, \method{} retains
\begin{equation}
    \widehat{k}_t=W_Ku_t+b_K,\quad z_{K,t}=A_Ku_t,\qquad
    \bar v_t=W_Vu_t+b_V,\quad z_{V,t}=A_Vu_t.
    \label{eq:transition-state}
\end{equation}
Here $\widehat{k}_t$ is the base key projection before RoPE. For any candidate version $v'$, migration computes
\begin{align}
    k_t^{(v')}&=R_t\!\left(\widehat{k}_t+s_KB_K^{(v')}z_{K,t}\right),
    \label{eq:key-migration}\\
    v_t^{(v')}&=\bar v_t+s_VB_V^{(v')}z_{V,t}.
    \label{eq:value-migration}
\end{align}
All historical final-block K/V tensors are reconstructed by batched matrix products over retained coordinates. Reconstruction starts from immutable base state, avoiding error accumulation from successive parameter deltas. Lower-block caches are unchanged. After migration, the runtime reevaluates block $L$ at $t=n$ under the candidate final-block Q/O/MLP adapters and migrated cache to obtain candidate logits. Earlier final-block outputs are unnecessary for future autoregressive steps.

Fixing $A$ enables compact retention. Exact migration also supports trainable down-projections by retaining each $u_t$ alongside the immutable base projections and recomputing candidate coordinates at publication. An optional \emph{normalized-input} representation retains $u_t$ alone and recomputes both base and adapter projections. Both preserve the invariant boundary at a higher memory or publication cost (Table~\ref{tab:retention}). If an earlier block is adapted, its frozen input can instead be retained and the entire mutable suffix replayed (Appendix~\ref{app:proof}).

\begin{figure}[t]
    \centering
    \includegraphics[width=\linewidth]{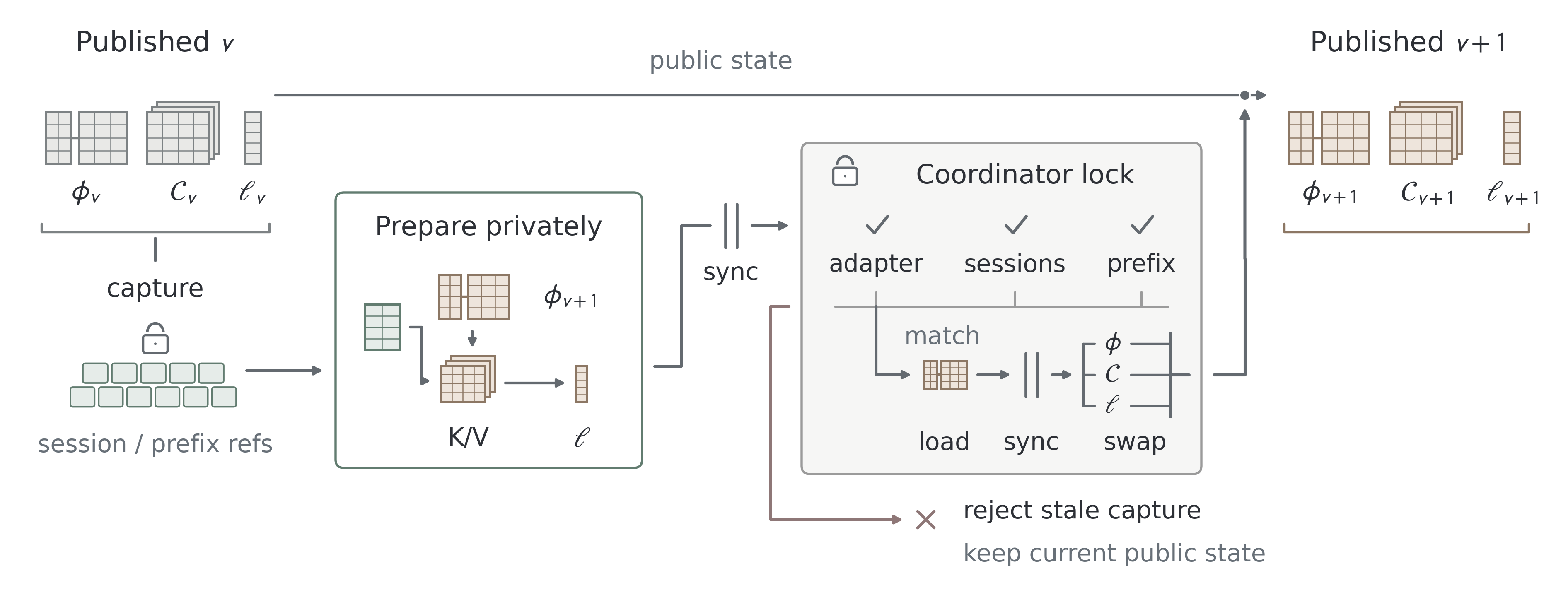}
    \caption{Coherent adapter publication.
    K/V reconstruction and latest-token reevaluation run privately.
    The coordinator validates captured references and publishes the adapter, cache, and logits together; stale captures are rejected.}
    \label{fig:publication}
\end{figure}

\begin{table}[h]
\caption{Retention choices for exact publication. Counts are dominant per-session history terms for $n$ tokens and $b$ bytes per scalar; the conventional KV cache, metadata, and latest-token state are additional. The trainable-$A$ ablation uses the original publication protocol (Appendix~\ref{app:memory}).}
\label{tab:retention}
\centering
\small
\setlength{\tabcolsep}{4pt}
\begin{tabular}{@{}p{0.23\linewidth}p{0.30\linewidth}p{0.40\linewidth}@{}}
\toprule
Representation & Retained history bytes & Candidate-version work \\
\midrule
Compact, fixed $A$ & $bn(d_K+d_V+r_K+r_V)$ & Reconstruct K/V; reevaluate latest token \\
Trainable $A$ & $bn(d_K+d_V+d)$ & Recompute coordinates and K/V; reevaluate latest token \\
Normalized inputs & $bnd$ & Recompute base and adapter K/V; reevaluate latest token \\
Boundary residuals & $bnd$ & Replay the final block over the history \\
\bottomrule
\end{tabular}
\end{table}

\subsection{Publication protocol}

Training uses a separate adapter copy. Initialization holds the serving coordinator lock while copying parameters and caching activations at the frozen boundary; optimization runs after the lock is released. At publication, the runtime captures active session states and prepares candidate tensors without changing their public references. Preparation can occur outside the coordinator lock. After synchronizing that work, commit reacquires the lock, verifies the adapter, session set, and captured prefixes, loads the candidate, synchronizes again, and swaps all state references. The runtime rejects the prepared update if a prefix has changed. Serving operations use the same coordinator, and the evaluated default CUDA stream orders device work; no asynchronous-overlap benefit is assumed. An exception before commit preserves the old published version. The guarantee is process-local failure atomicity.

\subsection{Correctness and cost}

\begin{theorem}[Conditional continuation equivalence]
\label{thm:exact}
Fix $x_{1:n}$, its mask, and position metadata. Assume that per-layer KV caches are the model's only cross-token hidden state; evaluation is deterministic; embeddings, computation through block $L-1$, $N_L$, the final normalization, and the language-model head are unchanged; final-block K/V changes only as in Eqs.~\eqref{eq:key-version}--\eqref{eq:value-version}, with fixed $A_K,A_V$ and linear $R_t$; and all other version-dependent final-block operations are reevaluated at the latest and future tokens. In exact arithmetic, Eqs.~\eqref{eq:key-migration}--\eqref{eq:value-migration} and latest-token reevaluation produce the same next-token distribution as complete prefill under $\phi_{v+1}$. Conditional distributions remain equal for every fixed future continuation whose mask and position metadata are extended identically.
\end{theorem}

The result follows because the frozen lower blocks produce the same $u_t$, so Eqs.~\eqref{eq:key-migration}--\eqref{eq:value-migration} reproduce every candidate-version final-block key and value; the latest-token final-block computation then receives the same inputs and parameters as full prefill. Induction gives the continuation statement. Appendix~\ref{app:proof} gives the proof. The theorem concerns exact arithmetic and this restricted adapter family, not bitwise equality between floating-point kernels.

For one $n$-token session, K/V rematerialization costs
\begin{equation}
    \Theta\!\left(nr_Kd_K+nr_Vd_V\right),
    \label{eq:migration-cost}
\end{equation}
plus one final-block attention scan and adapter loading. Across sessions, $n$ is replaced by total retained tokens. Dominant persistent migration state is $bn(d_K+d_V+r_K+r_V)$ bytes for $b$-byte scalars, in addition to the standard KV cache; per-session metadata, the latest-token residual and logits, and a serving-adapter snapshot are also retained. Complete accounting appears in Appendix~\ref{app:storage}.

\section{Related work}
\label{sec:related}

\paragraph{Fixed bases and factorized adapter caches.}
LoRA-FA fixes the LoRA down-projection $A$ to reduce training activation memory \citep{zhang2023lorafa}; \method{} repurposes fixed $A$ as version-invariant coordinates for historical K/V. LRAgent and ForkKV factor shared base and adapter-dependent cache state for multi-LoRA agents processing common contexts \citep{jeon2026lragent,wang2026forkkv}. They target reuse across adapter identities; \method{} instead migrates one active session across successive versions of the same adapter.

\paragraph{Restoration and version policy.}
HCache restores KV from stored intermediate activations, whereas DroidSpeak selectively recomputes layers when sharing caches across fine-tuned or updated models of the same architecture \citep{gao2025hcache,liu2026droidspeak}. LOCAL associates cache entries with adapter versions, invalidates stale entries, and schedules prefill before publication \citep{liu2026local}. \method{} instead uses an invariant final-block boundary for closed-form K/V rematerialization and coherent publication of the adapter, cache, and next-token logits.

\paragraph{Other cache-reuse and serving systems.}
Punica, S-LoRA, and SGLang optimize adapter batching, paging, or prefix reuse; Activated LoRA, ICaRus, and PrefillShare preserve reusable state through activation or training constraints; and MobiLoRA and cross-model mappings compress or transfer cache state \citep{chen2024punica,sheng2024slora,zheng2024sglang,greenewald2025activated,woo2026icarus,woo2026prefillshare,li2025mobilora,heo2026crossmodel}. Dynamic evaluation and CoLLM combine online adaptation with streaming inference or continuous serving, but do not directly study migration of an active cache across published adapter versions \citep{rannentriki2024dynamic,huang2026collm}. These mechanisms are complementary to \method{}'s publication transition.

\section{Evaluation}
\label{sec:evaluation}

We evaluate Qwen2.5-3B-Instruct \citep{qwen2024qwen25} on one NVIDIA H200 GPU. Unless otherwise noted, final-block Q/O/MLP adapters have rank 8 and $\alpha=16$; fixed-$A$ K/V adapters have rank 32 and scale 1. Gaussian adapter updates and controlled histories isolate state migration from the learning rule. Our evaluation measures migration fidelity and systems cost; task-level adaptation quality under the final-block and fixed K/V down-projection restrictions remains unevaluated. We distinguish three protocols, each with its own implementation and timing scope.

\paragraph{Publication scaling.}
The original study uses five seeds, fixed random-token histories, 30 paired updates per ordinary run, and the original latest-token kernel; stress runs use 100 updates. Timing spans candidate availability to next-token logits. Migration includes its locked commit, while full re-prefill excludes reference-state installation. We compute 95\% confidence intervals (CIs) using a two-stage hierarchical bootstrap for this protocol (Appendix~\ref{app:systems-protocol}).

\paragraph{Matched growing-history serving.}
The compact-retention runtime uses the model's native scaled dot-product attention (SDPA) implementation, immutable history chunks, and common reference-selected continuation tokens. Migration, retained-boundary replay, and full-history re-prefill use the same publication coordinator, guards, and commit semantics. Boundary replay retains historical final-block residuals and shares lower-block K/V. Decoding benchmarks use three seeds in bfloat16 (BF16) at initial context lengths of 2,048, 8,192, and 32,000 tokens, with eight warmup and 48 measured tokens per case. Separate 8K traces use 32 or 128 tokens between updates, one warmup publication, and two measured publications. These intervals match token increment sizes studied in streaming dynamic evaluation \citep{rannentriki2024dynamic}; they are sensitivity points, not estimates of deployed agent publication frequency.

The guarded direct-serving baseline includes coordinator ownership checks, prefix and mask retention, cache metadata, and commit. The native cached-decoding baseline omits these runtime services. We time synchronized complete operations, with numerical diagnostics excluded from timing. Publication-plus-decoding totals exclude initial prefill and training; all alternatives receive the same candidate adapter. We average costs within each seed and then average the seed means. Three-seed ranges are descriptive, not confidence intervals. Appendix~\ref{app:revised} records correctness coverage and precision controls.

\paragraph{Retention alternatives.}
A separate three-seed normalized-input trace measures four publications, each followed by 256 tokens, after eight warmup tokens at an 8K prefix. It compares migration and boundary replay under the same coordinator. This protocol quantifies an alternative storage representation; its timings are not pooled with the compact-retention study (Appendix~\ref{app:normalized}).

\section{Results}
\label{sec:results}

\begin{figure}[t]
    \centering
    \includegraphics[width=\linewidth]{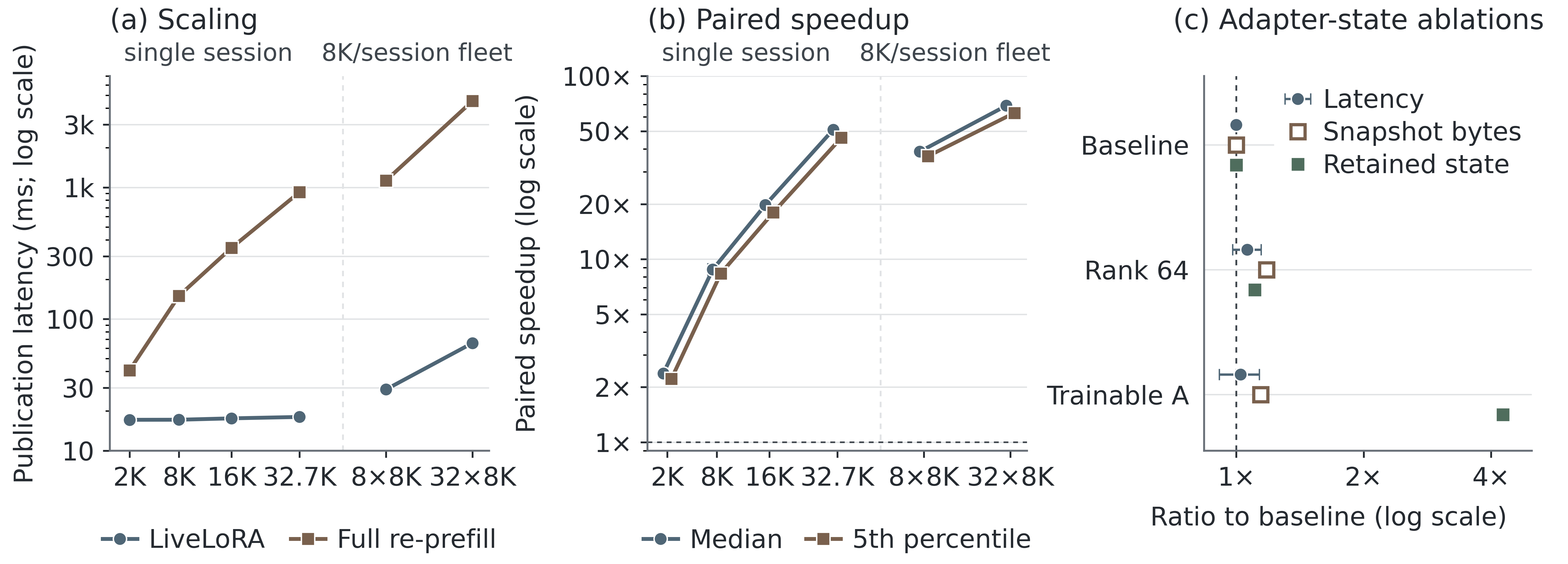}
    \caption{Original five-seed publication study. (a) Mean within-run median publication latency. (b) Mean within-run median and 5th-percentile paired re-prefill/migration speedups. (c) Publication, snapshot, and retained-state ratios relative to fixed-$A$ rank 32 at 8K; trainable-$A$ retains normalized inputs. Error bars are two-stage hierarchical-bootstrap 95\% intervals over five 30-update runs; byte ratios are exact tensor counts. This study uses the original latest-token kernel and publication timer scope described in Section~\ref{sec:evaluation}.}
    \label{fig:scaling}
\end{figure}

\paragraph{Publication scales to long histories and session fleets.}
In the original study, single-session median migration latency changes from 17.15\,ms at 2K to 18.03\,ms at 32.7K, while full re-prefill grows from 40.78 to 919.21\,ms. The mean within-run median paired speedup reaches $51.06\times$ [50.10, 52.07] at 32.7K. For 32 sessions with 8K histories it reaches $68.98\times$ [67.08, 71.39], with a within-run 5th-percentile speedup of $62.96\times$ [60.61, 66.06] (Figure~\ref{fig:scaling}). These results characterize publication, separately from the matched serving costs below.

\begin{table}[t]
\caption{Compact-retention decoding with native SDPA. Values are means across three seeds, each with 48 measured tokens per initial context; BF16, one session. Native cached decoding is a diagnostic baseline. Guarded direct serving includes the same coordinator safeguards as \method{}.}
\label{tab:decode}
\centering
\small
\setlength{\tabcolsep}{5pt}
\begin{tabular}{@{}lrrrrr@{}}
\toprule
 & \multicolumn{3}{c}{Latency (ms/token)} & \multicolumn{2}{c}{Added cost versus} \\
\cmidrule(lr){2-4}\cmidrule(l){5-6}
Context & Native & Guarded & \method{} & Native & Guarded \\
\midrule
2K & 20.851 & 21.699 & 21.855 & 4.81\% & 0.72\% \\
8K & 20.829 & 21.642 & 21.818 & 4.75\% & 0.81\% \\
32K & 21.604 & 22.506 & 22.659 & 4.88\% & 0.68\% \\
\bottomrule
\end{tabular}
\end{table}

\paragraph{Capturing migration state adds modest decoding cost.}
Reusing the model's normalization, base projections, and low-rank coordinates avoids duplicate projection work. Retained tensors are stored in immutable chunks and concatenated only at publication, avoiding full-history copies at each decoding step. The resulting compact runtime adds 4.75--4.88\% over native cached decoding and 0.68--0.81\% over equally guarded direct serving, including ownership checks (Table~\ref{tab:decode}).

\begin{table}[t]
\caption{Matched compact-retention publication and serving at 8K. Publication columns are mean milliseconds per update. Total columns sum two measured publications and 64 or 256 generated tokens; three equally weighted seeds in BF16. All methods include coordinator safeguards and commit. Initial prefill, training, warmups, and diagnostics are outside these totals.}
\label{tab:trace}
\centering
\small
\setlength{\tabcolsep}{4pt}
\begin{tabular}{@{}lrrrrrr@{}}
\toprule
 & \multicolumn{3}{c}{Publication (ms/update)} & \multicolumn{3}{c}{Publication + decoding (ms)} \\
\cmidrule(lr){2-4}\cmidrule(l){5-7}
Tokens/update & \method{} & Boundary & Full & \method{} & Boundary & Full \\
\midrule
32 & 9.66 & 15.11 & 155.82 & 1,429.29 & 1,429.25 & 1,706.36 \\
128 & 10.55 & 15.47 & 156.87 & 5,662.15 & 5,628.49 & 5,892.03 \\
\bottomrule
\end{tabular}
\end{table}

\paragraph{Retention determines the publication--serving tradeoff.}
The matched traces save 16.24\% and 3.90\% of measured publication-plus-decoding time versus full-history re-prefill at cadences 32 and 128 (Table~\ref{tab:trace}). Compared with retained-boundary replay, compact migration reduces additional session storage by $3.42\times$: 9.48\,MiB versus 32.41\,MiB at the initial 8K prefix. Mean publication-plus-decoding costs remain within 0.60\% of boundary replay across both cadences. These logical storage counts exclude shared model/coordinator storage and publication workspaces.

Normalized-input retention offers a different point on this tradeoff. In its separate cadence-256 study, mean publication latency is 6.55\,ms versus 15.95\,ms for boundary replay, while publication-plus-decoding totals are effectively tied. It retains approximately as much state as boundary replay (Appendix~\ref{app:normalized}).

\paragraph{Finite-precision checks.}
In the decoding microbenchmarks, \method{} and guarded direct serving show no observed differences in logits or checked cache tensors. Across repeated BF16 publications, migration and boundary replay agree on the tested greedy token selections, with nonzero differences in final-block K/V and logits. Dense re-prefill and ordinary cached execution also differ numerically. Appendix~\ref{app:revised} reports the references, error magnitudes, and 32-bit floating-point (FP32) controls.

\paragraph{Choosing a publication policy.}
For $N$ tokens between supplied candidate updates, a local cost model is
\begin{equation}
    C_j(N)=P_j+Nd_j,\qquad
    P_R-P_M>N(d_M-d_R),
    \label{eq:interval-cost}
\end{equation}
where $P_j$ is publication cost and $d_j$ is per-token serving cost. The inequality gives the condition for migration $M$ to amortize against replay $R$. Using publication and decoding costs from Table~\ref{tab:trace}, the local break-even intervals at 8K are approximately 600 tokens per update against full-history re-prefill and 30 against retained-boundary replay. These local model estimates assume constant publication and per-token serving costs. Training costs cancel for publication mechanisms receiving the same candidate, but must be included when comparing learning policies or complete application lifetimes. Session length, available memory, and publication-latency requirements also inform strategy selection.

\section{Scope and future work}

The compact representation targets final-block adaptation with fixed K/V adapter down-projections. Normalized-input retention accommodates trainable down-projections, while adapting earlier blocks can be supported by replaying the mutable suffix (Appendix~\ref{app:proof}). The implementation uses contiguous sequences, standard RoPE, and process-local coordination. Future studies can pair learned adapters with application-driven update schedules to assess adaptation quality alongside publication latency and total serving cost. Integrating the publication contract with paged caches and distributed serving would broaden its use in existing inference systems.

\section{Conclusion}

\method{} turns a frozen activation boundary into a publication primitive for adapter parameters, KV state, and next-token logits. Compact coordinates reduce the state retained for migration, while normalized-input retention and suffix replay expose alternative memory and recomputation choices. The experiments establish substantial publication savings over full-history re-prefill and quantify the smaller tradeoff against retained-boundary replay. By publishing parameters together with the inference state needed to execute them, \method{} makes adapter updates coherent transitions of persistent agent state.

\bibliographystyle{plainnat}
\bibliography{references}

\appendix

\section{Proof of Theorem~\ref{thm:exact}}
\label{app:proof}

Let complete prefill under the candidate version be the reference execution. Both executions use the same mask, position metadata, normalization, and positional map. Because embeddings and blocks $1{:}L-1$ are fixed and evaluation is deterministic, they produce identical lower-block caches and final-block inputs $u_t$ for every $t\le n$. The retained state satisfies
\begin{equation}
\widehat{k}_t=W_Ku_t+b_K,\quad z_{K,t}=A_Ku_t,\quad
\bar v_t=W_Vu_t+b_V,\quad z_{V,t}=A_Vu_t.
\end{equation}
Migration therefore gives
\begin{align}
R_t\!\left(\widehat{k}_t+s_KB_K^{(v+1)}z_{K,t}\right)
&=R_t\!\left(W_Ku_t+b_K+s_KB_K^{(v+1)}A_Ku_t\right),\\
\bar v_t+s_VB_V^{(v+1)}z_{V,t}
&=W_Vu_t+b_V+s_VB_V^{(v+1)}A_Vu_t.
\end{align}
These are exactly the final-block keys and values produced by candidate-version prefill at every historical position. Reevaluation at $t=n$ receives the same pre-normalization residual, query, K/V tensors, mutable final-block parameters, final normalization, and language-model head as the reference, and therefore produces identical logits in exact arithmetic.

For induction, assume both executions have consumed the same arbitrary continuation through position $t\ge n$, extended mask and position metadata identically, and have equal caches. At $t+1$, fixed lower blocks receive the same token and cache and yield the same final-block input. The candidate final block then appends the same K/V and produces the same logits. Conditional distributions are therefore equal for every fixed continuation. Deterministic decoding gives the same path; under sampling, equality follows when random variates are coupled. \hfill$\blacksquare$

\paragraph{Trainable down-projections.}
When the immutable state retains $u_t$, candidate coordinates can be computed as $z_{K,t}^{(v')}=A_K^{(v')}u_t$ and $z_{V,t}^{(v')}=A_V^{(v')}u_t$. Substituting these into Eqs.~\eqref{eq:key-migration}--\eqref{eq:value-migration} reproduces candidate-version K/V by the same algebra. Latest-token reevaluation and the continuation induction are unchanged. Fixed $A$ therefore determines which historical state is retained, while the frozen lower computation supplies the invariant boundary.

\paragraph{Multiple adapted blocks.}
If block $j<L$ is the earliest adapted block, the residual $h_{j-1,t}$ remains invariant, but higher-block inputs generally do not. Retaining all historical boundary residuals permits exact replay of blocks $j{:}L$ under the candidate version, rebuilding their K/V caches with the same masks and positions. This avoids rerunning the frozen lower blocks but requires history-wide attention and feed-forward computation in the adapted suffix, together with $bnd$ bytes of boundary residuals for one $n$-token session. Fixed down-projections alone do not eliminate this dependence. This extension describes a correctness-preserving construction; its systems performance is not established by the final-block experiments.

\section{Implementation and systems protocol}
\label{app:systems-protocol}

\paragraph{Model and adapter.}
We use Qwen2.5-3B-Instruct with grouped-query attention, so $d_K,d_V$ denote aggregate KV-head widths. The main configuration uses rank-8 final-block Q/O/MLP LoRA with $\alpha=16$ and orthogonally initialized rank-32 fixed-$A$ final-block K/V LoRA with $\alpha_{KV}=32$, so $s_K=s_V=1$. Each session accepts one unpadded sequence with an all-ones mask and contiguous zero-based positions; fleet points use multiple session objects, while reference re-prefill uses one dense, equal-length batch. Sliding-window attention and RoPE scaling are disabled. Inference is in evaluation mode on the default CUDA stream.

\paragraph{Publication implementation.}
Candidate cloning and validation precede preparation. For runtime-managed sessions using analytic migration, the coordinator captures the adapter, session set, and immutable session-state references under its lock, then prepares candidate K/V and logits outside it. After device synchronization, commit reacquires the lock, verifies those references, loads the candidate, synchronizes, and publishes the adapter and prepared states. If a serving step changed a captured prefix, publication is rejected. Replay strategies perform candidate loading, reconstruction, synchronization, and commit under the same coordinator lock. Private preparation leaves published state unchanged. If replay preparation fails under the coordinator lock, the prior adapter is restored and session references are preserved. These are process-local precommit guarantees; durable recovery and distributed publication are outside the implementation.

Initializing the training copy acquires the coordinator lock while copying the adapter or adapted suffix and collecting training inputs at the frozen boundary. Optimization uses this independent copy after lock release. Training can be scheduled between serving operations. All CUDA work uses the default stream, and the reported timings assume serial execution.

\paragraph{Runtime and statistics.}
The original fixed-history publication, re-prefill, and correctness measurements use BF16 model and cache tensors with SDPA, except for the custom latest-token kernel described below. Ordinary runs use five warmup iterations and 30 measured paired updates; stress runs use 100 measured updates. Each case has five independent seeds. Equal-tailed 95\% CIs are computed from 10,000 two-stage hierarchical bootstrap draws that resample runs and then paired observations within sampled runs.

\paragraph{Original and revised attention paths.}
The original fixed-history publication measurements and publication-time numerical comparisons use a custom latest-token kernel, which uses matrix multiplication with FP32 softmax. The revised runtime dispatches latest-token attention through the model's native eager or SDPA implementation, with one query attending all retained positions. This implementation change does not alter the migration equations or the exact-arithmetic theorem. The growing-history measurements in Section~\ref{sec:results} evaluate the revised path separately. The original continuation comparisons evaluate the same reference-selected token from migrated and reference caches using SDPA in both executions. Tables~\ref{tab:decode}--\ref{tab:trace} report matched publication and decoding measurements using the revised runtime; Appendix~\ref{app:normalized} reports normalized-input retention.

\section{Original fixed-history systems study}
\label{app:systems-results}

\paragraph{Migrated K/V tensors match reference re-prefill; the custom latest-token kernel differs numerically.}
Across all seven cases, migrated final-block keys and values have zero observed maximum absolute error against independently captured tensors from reference re-prefill. The custom latest-token kernel produces next-token logit mean absolute error (MAE) of 0.0168--0.0201 and 96.0--97.9\% top-1 agreement in the six scaling cases; under the 100-update stress condition, aggregate MAE is 0.0722 and top-1 agreement is 93.8\%. Because the compared final-block K/V tensors match exactly in these runs, this discrepancy is localized to the custom kernel and is consistent with a different floating-point execution order. To isolate cache reconstruction, we evaluate the same reference-selected token from migrated and reference caches using SDPA in both executions. The resulting logits have zero observed MAE and 100\% top-1 agreement in every comparison. By contrast, stale K/V causes 8.0--17.3\% top-1 disagreement in the six scaling cases and 35.4\% [30.0, 40.6] under the stress condition. Figure~\ref{fig:coherence} in the appendix separates these sources of discrepancy.

\begin{table}[ht]
\caption{Publication scaling in the original fixed-history study. Latencies average within-run medians; speedups average within-run medians and 5th percentiles of paired re-prefill/\method{} ratios. Brackets are two-stage hierarchical-bootstrap 95\% CIs. Each ordinary case uses five 30-update runs; $^{\dagger}$ marks five 100-update runs.}

\label{tab:systems-full}
\centering
\small
\resizebox{\linewidth}{!}{%
\begin{tabular}{@{}rrrrrr@{}}
\toprule
Tokens & Sessions & LiveLoRA ms & Re-prefill ms & Median speedup & 5th-pctl. speedup \\
\midrule
2,048 & 1 & 17.15 [16.31,17.93] & 40.78 [40.40,41.10] & 2.38 [2.29,2.48] & 2.22 [2.19,2.27] \\
8,192 & 1 & 17.17 [15.94,18.29] & 150.05 [149.40,150.69] & 8.79 [8.23,9.45] & 8.36 [7.89,8.90] \\
16,384 & 1 & 17.59 [16.75,18.20] & 347.19 [346.00,348.36] & 19.78 [19.10,20.77] & 18.01 [17.36,18.60] \\
32,700 & 1 & 18.03 [17.68,18.34] & 919.21 [914.45,923.63] & 51.06 [50.10,52.07] & 46.09 [44.81,47.80] \\
8,192 & 8 & 29.16 [28.63,29.63] & 1,128.21 [1,121.70,1,135.47] & 38.70 [38.10,39.39] & 36.51 [35.60,37.12] \\
8,192 & 32 & 65.71 [63.80,67.60] & 4,534.46 [4,511.63,4,557.90] & 68.98 [67.08,71.39] & 62.96 [60.61,66.06] \\
8,192$^{\dagger}$ & 1 & 17.34 [15.73,18.82] & 149.81 [149.20,150.51] & 8.75 [8.00,9.63] & 7.52 [6.93,8.01] \\
\bottomrule
\end{tabular}
}
\end{table}

\section{Correctness and stale-cache control}
\label{app:correctness}

Stale-cache Kullback--Leibler divergence, $\KL(p_{\mathrm{ref}}\|p_{\mathrm{stale}})$, compares matched-SDPA outputs after the same reference-selected token under candidate weights. Here $p_{\mathrm{ref}}$ uses full-history reference K/V and $p_{\mathrm{stale}}$ uses prior-version K/V (Appendix~\ref{app:systems-results}). Across the six scaling points it ranges from 0.0272 to 0.0375; under the 100-update stress condition it is 0.4864 [0.3494,0.6341]. The stale-cache control is intentionally invalid and is not a deployable latency baseline.

\begin{figure}[t]
\centering
\includegraphics[width=\linewidth]{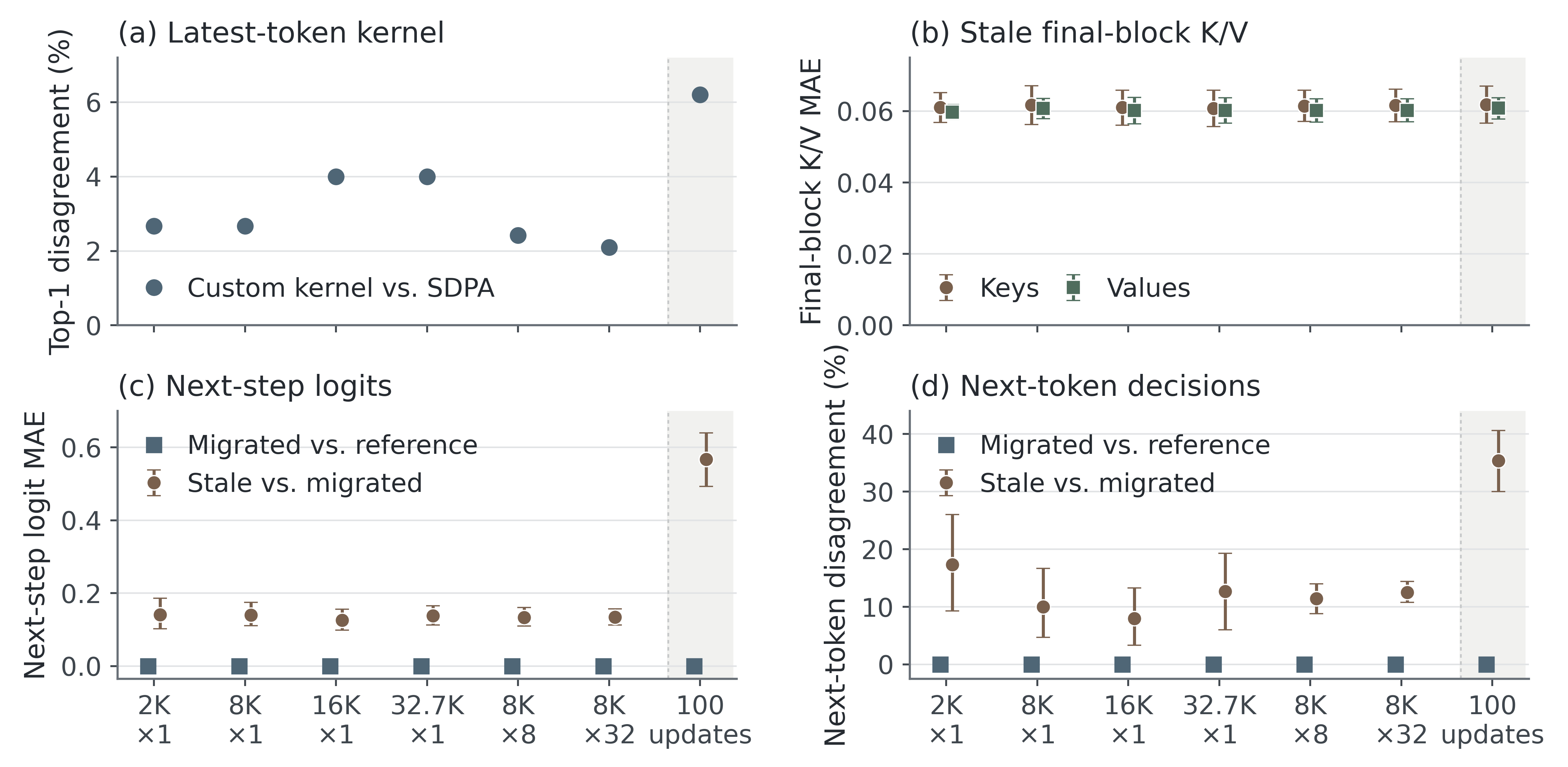}
\caption{Cache-coherence diagnostics. (a) Custom-kernel versus reference top-1 disagreement. (b) Stale versus migrated final-block K/V MAE. (c,d) Logit MAE and top-1 disagreement after a common reference-selected token using SDPA; migrated and reference outputs coincide in these tests. Stale-state error bars are two-stage hierarchical-bootstrap 95\% CIs. Shading marks the 100-update stress case; ordinary cases use 30 updates per run (five runs each).}
\label{fig:coherence}
\end{figure}

\section{Memory, adapter-state ablations, and accounting}
\label{app:memory}
\label{app:storage}

\begin{table}[ht]
\caption{Memory and decoding overhead in the original fixed-history study. Analytical memory excludes training state, workspaces, and caller-owned candidates; percentages use unrounded byte counts. Decode overhead for migration-state capture averages within-run median percentage overheads over 20 cached-decoding steps relative to direct cached evaluation. Brackets are two-stage hierarchical-bootstrap 95\% CIs.}
\label{tab:memory-full}
\centering
\small
\setlength{\tabcolsep}{4pt}
\begin{tabular}{@{}lrrrr@{}}
\toprule
Case & KV cache MiB & LiveLoRA MiB & Steady OH & Decode OH [95\% CI] \\
\midrule
2K $\times$ 1 & 72.0 & 76.3 & 6.04\% & 21.94\% [20.34,23.38] \\
8K $\times$ 1 & 288.0 & 299.2 & 3.90\% & 22.33\% [20.96,23.62] \\
16K $\times$ 1 & 576.0 & 596.4 & 3.55\% & 22.25\% [21.31,23.40] \\
32.7K $\times$ 1 & 1,149.6 & 1,188.3 & 3.37\% & 19.17\% [18.13,20.11] \\
8K $\times$ 8 & 2,304.0 & 2,381.6 & 3.37\% & 24.24\% [22.81,25.51] \\
8K $\times$ 32 & 9,216.0 & 9,521.2 & 3.31\% & 24.02\% [22.46,25.51] \\
100-update stress & 288.0 & 299.2 & 3.90\% & 23.72\% [22.39,24.93] \\
\bottomrule
\end{tabular}
\end{table}

The trainable-$A$ control uses standard LoRA with both factors trainable: it retains $u_t$, recomputes $A^{(v')}u_t$, and then applies $B^{(v')}$.

At 8K tokens and one session, the trainable-$A$, rank-32 ablation has publication-latency ratio 1.023 [0.912,1.133] relative to fixed-$A$ rank 32, retained migration-state ratio $4.270\times$, serving-adapter snapshot ratio $1.142\times$, and 10.851 percentage points more steady-state overhead. Fixed-$A$ rank 64 has publication-latency ratio 1.061 [0.980,1.145], retained migration-state ratio $1.105\times$, serving-adapter snapshot ratio $1.178\times$, and 0.456 percentage points more steady-state overhead. Both publication-latency ratio intervals include one.

For sessions $s=1,\ldots,m$ with lengths $n_s$, the dominant fixed-$A$ migration-state size and reported steady-state size are
\begin{align}
M_{\mathrm{mig}}&=b\sum_s n_s(d_K+d_V+r_K+r_V),\\
M_{\mathrm{steady}}&=M_{\mathrm{KV}}+M_{\mathrm{mig}}+M_{\mathrm{desc}}+M_{\mathrm{res}}+M_{\mathrm{logits}}+M_{\mathrm{snapshot}}.
\end{align}
Here $M_{\mathrm{KV}}$ is the standard KV cache; $M_{\mathrm{desc}}$ is per-session metadata; $M_{\mathrm{res}}$ and $M_{\mathrm{logits}}$ are latest-token residuals and next-token logits; and $M_{\mathrm{snapshot}}$ is the serving-adapter snapshot. When both $A$ and $B$ are trainable, retaining the normalized final-block input at each historical position replaces the coordinate term $b\sum_s n_s(r_K+r_V)$ with $b\sum_s n_sd$. Atomic publication additionally needs a second final-block cache, a snapshot of the candidate's trainable parameters, and candidate logits; this lower bound excludes operator workspaces, allocator effects, caller-owned candidate tensors, and training state, so it is not a measured peak.

\section{Matched serving references and precision controls}
\label{app:revised}

The matched compact-retention protocol uses three seeds and separate but identically initialized models for each strategy. All receive the same nonzero Gaussian adapter and reference-selected continuation tokens. Timed publication includes candidate cloning, ownership validation, preparation, synchronization, and commit. Timed decoding uses each session's normal step. Method order follows a balanced seed-dependent schedule. Two diagnostic paths measure native cached decoding and cached decoding with state capture but without history append or commit. Guarded direct serving and \method{} decoding include the complete coordinator operations.

\paragraph{Coverage.}
We evaluate decoding at three context lengths across three seeds, with eight warmup and 48 measured tokens per case. We compare logits and greedy decisions after every step, and full caches at initial and final prefixes; the compared paths show zero observed numerical differences. These checks occur outside timing. The six publication cases cover 12 measured publications and 960 measured continuation comparisons per strategy. Including warmups, correctness coverage is 18 publications and 1,440 continuations. Full-cache comparisons occur at every publication and interval endpoint; continuation comparisons evaluate logits, versions, and prefixes.

\paragraph{Independent references.}
At each publication, an independent dense prefill evaluates the updated model on the same history. Its cache also supplies a common-token cached continuation reference. A fresh dense prefill at each interval endpoint separately measures the difference between ordinary cached serving and dense execution. Retained-boundary replay shares the frozen lower-block history and independently reconstructs the candidate final block; it is a complementary reference rather than a replacement for dense prefill.

In BF16, migration and boundary replay agree on all tested greedy token selections and lower-block K/V. Maximum final-block key and value errors at publication are both 0.03125; maximum logit errors are 0.125 at publication and 0.25 during continuation. Against cached continuation from full-history re-prefill, migration's maximum logit difference is 1.1875 at cadence 32 and 5.59375 at cadence 128, again without an observed greedy disagreement. Ordinary cached-versus-fresh-dense checks also have nonzero differences. These observations characterize the evaluated floating-point executions; they do not assert equal distributions or universal greedy agreement.

\paragraph{Precision control.}
A separate earlier native-SDPA FP32 control uses one 8K history, three publications, and 24 common continuation tokens. Maximum direct migration--boundary logit errors are $2.43\times10^{-5}$ at publication and $7.63\times10^{-6}$ during continuation; final-block key/value errors are at most $1.24\times10^{-5}/1.10\times10^{-5}$. All compared greedy token selections agree. This control isolates precision sensitivity and is not pooled with the current BF16 performance protocol.

\paragraph{Interpretation and storage.}
Tables~\ref{tab:decode}--\ref{tab:trace} characterize a fixed runtime revision; host and process variation across sequential runs prevents attributing absolute latency differences to a single code change. Initial prefill is separately recorded and is not part of publication-plus-decoding totals. At the initial 8K prefix, compact and boundary session tensors occupy 9.481 and 32.415\,MiB, respectively. At the final cadence-128 prefix of 8,576 tokens they occupy 9.912 and 33.921\,MiB. Joining compact histories during publication needs an additional 9.422\,MiB materialization workspace at that endpoint. These are logical tensor counts, not allocator peaks. Metadata grows with the number of stored chunks.

\section{Normalized-input retention}
\label{app:normalized}

This optional final-block representation stores $u_t$ instead of historical base K/V and fixed-$A$ coordinates. At publication it recomputes frozen base and candidate adapter projections and then performs latest-token reevaluation using the model's native attention implementation. Its dominant history storage is $bnd$ bytes, plus token/mask metadata, latest residual, and logits. Unlike the trainable-$A$ representation in Appendix~\ref{app:memory}, it does not also retain the base K/V projections. The experiments use Qwen's native forward implementation. Both representations use candidate reconstruction rather than accumulated cache deltas.

The normalized-input experiments use three seeds in BF16, a constructed tool transcript of 8,192 tokens, eight warmup tokens, and four publications with 256 common continuation tokens after each. Both strategies use the same coordination protocol, adapter updates, and continuation tokens. Table~\ref{tab:normalized} gives means across seeds and descriptive seed ranges. The normalized-input and compact experiments use separate runtime revisions and GPU allocations, so their absolute latencies are not paired comparisons between retention modes.

\begin{table}[ht]
\caption{Normalized-input retention at cadence 256; three seeds. Each total sums four publications and 1,024 decoded tokens. Brackets are seed ranges. Initial prefill, training, warmups, and diagnostics are excluded.}
\label{tab:normalized}
\centering
\small
\setlength{\tabcolsep}{5pt}
\begin{tabular}{@{}lrr@{}}
\toprule
 & \method{}, normalized & Retained boundary \\
\midrule
Publication (ms/update) & 6.55 [5.65, 8.29] & 15.95 [14.95, 17.09] \\
Decoding (ms/token) & 19.45 [18.34, 20.90] & 19.42 [18.38, 20.80] \\
Publication + decoding (ms) & 19,940.61 & 19,945.65 \\
Final retained state (MiB) & 36.47 & 36.46 \\
\bottomrule
\end{tabular}
\end{table}

Shorter publication latency accompanies effectively tied publication-plus-decoding cost: mean paired saving is 5.05\,ms out of approximately 19.95\,s, with a seed range of $-59.32$ to $58.36$\,ms. Two seeds favor migration and one favors boundary replay. An earlier normalized-input implementation likewise differs from boundary replay by less than 0.1\% in its cadence-256 and cadence-1024 totals.

At the final 9,224-token prefix, normalized history occupies 36.03\,MiB. Total additional session state is 36.47\,MiB versus 36.46\,MiB for boundary replay, with 4,104 extra bytes for migration's latest residual and position. Materialization requires another 36.03\,MiB, separately from these persistent counts. Allocator, kernel, optimizer, and Python overheads are excluded.

Across the three cadence-256 runs, 12 publication and 3,072 continuation comparisons agree on greedy token selections, versions, and prefixes. Full-cache comparisons at the initial and final prefixes (six in total) show identical lower-block K/V and maximum final-block K/V differences of 0.03125 relative to retained-boundary replay. The maximum logit difference across publication and continuation comparisons is 0.25.

\end{document}